# Automatic Analysis of Linguistic Features in Journal Articles of Different Academic Impacts with Feature Engineering Techniques

**Siyu Lei**
Xi'an Jiaotong University, Xi'an, China
The Hong Kong Polytechnic University, Hong Kong, China
siyu525.lei@connect.polyu.hk
**Ruiying Yang**[*]
Xi'an Jiaotong University, Xi'an, China
yangryd@xjtu.edu.cn

**Chu-Ren Huang**[**]
The Hong Kong Polytechnic University, Hong Kong, China
The Hong Kong Polytechnic University-Peking University Research Centre on Chinese Linguistics, Hong Kong, China
churen.huang@polyu.edu.hk

## Abstract

English research articles (RAs) are an essential genre in academia, so the attempts to employ NLP to assist the development of academic writing ability have received considerable attention in the last two decades. However, there has been no study employing feature engineering techniques to investigate the linguistic features of RAs of different academic impacts (i.e., the papers of high/moderate citation times published in the journals of high/moderate impact factors). This study attempts to extract micro-level linguistic features in high- and moderate-impact journal RAs, using feature engineering methods. We extracted 25 highly relevant features from the Corpus of English Journal Articles through feature selection methods. All papers in the corpus deal with COVID-19 medical empirical studies. The selected features were then validated of the classification performance in terms of consistency and accuracy through supervised machine learning methods. Results showed that 24 linguistic features such as the overlapping of content words between adjacent sentences, the use of third-person pronouns, auxiliary verbs, tense, emotional words provide consistent and accurate predictions for journal articles with different academic impacts. Lastly, the random forest model is shown to be the best model to fit the relationship between these 24 features and journal articles with high and moderate impacts. These findings can be used to inform academic writing courses and lay the foundation for developing automatic evaluation systems for L2 graduate students.

## 1 Introduction

Academic writing has been considered one of the most difficult and challenging parts for postgraduates with English as their second language. To improve the English learners' writing ability, many corpus studies compare L2 learners' writing with native experts' writing in terms of specific linguistic features and point out the progress that L2 learners should make (e.g., Lorenz, 1999; Hyland, 2008; Qin, 2014). Native experts' writing in the above studies usually includes high-impact English journal articles (i.e., those with high citations and published in the journals with high-impact factors) because they can be one of the indicators of high-quality academic products and thus become a good reference for assessing students' academic writing. However, native experts' publication has been proofread concerning linguistic use, while students' academic writings have more or less grammatical problems or spelling errors. Directly comparing such two types of writing is inevitably involved in those issues, but many studies do not mention such issues when introducing student corpora (e.g., Qin, 2014; Lu, 2012; Lu & Deng, 2019), which might cause the neglect of some word analysis. Nevertheless, comparison of publications with different impacts in academic fields can overcome such a problem, after all, both have been reviewed and proofread by experts. Grammatical problems might not be encountered to a bigger extent. At the same time, the most prominent weighted linguistic features in high-impact journal articles by comparison with those of moderate-impact can also provide a reference for the assessemtn of students' writings. Hence, it could be a new insight for L2 academic writing studies by comparing journal articles with high- and moderate-impacts.

As NLP technology has an increasingly close association with English writing classrooms, an increasing number of studies on linguistic use in English research articles introduce machine learning models for analysis. *Mover*, developed by Anthony and Lashkia (2003), maybe the earliest attempt in automatic analysis of English research article writing. It focuses on the move/step in the abstract sections of academic papers based on the Bag-of-Words model. *MAZEA* (Multi-label Argumentative Zoning for English Abstracts, Dayrell et al., 2012), also automatically analyzes the move/step of English academic papers' abstracts. The highest accuracy rate of the software is 69%. Based on SVM (support vector machine), *IADE* (Intelligent Academic Discourse Evaluator), and Research Writing Tutor (*RWT*) by the Cotos team of Iowa State

University in the United States (Pendar & Cotos, 2008; Babu, 2013; Cotos, 2014) vectorize the move/step in the Bag-of-Words Model. However, current automatic analysis systems on research articles focus more attention on the move/step at the macro-level and neglect the feature extraction at the micro-level of research articles. The importance of micro-level linguistic features have been proved by a series of corpus-based studies of second language acquisition (such as Biber et al., 2011; Biber & Gray, 2016; Chiu et al., 2017; Boutron & Ravaud, 2018; Lei & Yang, 2020; Politzer-Ahles et al., 2020). Liu (2016) and Wang and Liu (2017) also proved the importance of linguistic features at micro-levels in research articles. In their automatic classifier of abstracts in applied linguistics journals, classification accuracy (information abstract and descriptive abstract) has been significantly increased to 78.19% by introducing micro-level language indicators (such as sentence length, predicates, and connectives). The result is much higher than the accuracy rate of automatic analysis systems by Anthony and Lashkia (2003) and Dayrell et al. (2012). Hence, the micro-level linguistic features of high-impact English academic papers should be considered and analyzed to provide a greater number of dimensions for future automatic analysis and feedback systems of research articles.

Thanks to the rapid development of NLP technology, many emerging methods can help automatically extract features with bigger weights beyond corpus-based studies, for example, feature engineering. It has been widely applied in fault detection (e.g., Li et al., 2021), image detection (e.g., Cai, Nee, & Loh, 1996; Cai & Chen, 2011), etc., but has not been tried in the micro-level linguistic feature extraction of research articles. Our study will employ this method to extract and validate the selected features to provide consistent and accurate predictions for journal articles with different academic impacts.

## 2  Our Study

To sum up, the core issue of this study is to explore which micro-level linguistic features have a strong classificability on binary-impact journal articles and which machine learning model is better for fit the relationship between these micro-level linguistic indicators and the impact of journal articles. To control the influence of disciplines and themes on classification, we chose the papers in medical COVID-19 empirical. Specifically, we propose three research questions:

(1) By feature engineering techniques, which linguistic features have a more substantially explanatory effect on the binary-impact journal papers of COVID-19 medical empirical research?
(2) Can the machine learning model trained on the selected linguistic features consistently and accurately predict the impact of journal articles?
(3) Based on the performance of the above machine learning models, which machine learning model performs best?

By answering the above research questions, our study has two contributions to the field of NLP:

- ❖ The linguistic indicators extracted and verified in this study provide a more micro-level linguistic dimension for the future automatic analysis and feedback system of English research article writing, thereby improving the accuracy of evaluation and expanding the dimension of feedback.
- ❖ The classification models of journal articles with different impacts can be used as a reference for identifying and predicting the academic impact of manuscripts by naive writers and recommending target journals that match the writing quality of the manuscript.

## 3  Data, Features, and Method

### 3.1 Data Set Selection

The data sets of this study were selected from three databases, i.e., Web of Science, Scopus, and the Chinese Academy of Sciences Journal Division (hereafter CAS). We used the topic of "COVID-19" and "SARS-CoV-2" to search papers in the Web of Science and Scopus. The publication period was set within 2020, the document category is articles of empirical studies, and the language is written in English. The disciplines were set in the medical subjects, with the sub-subjects immunology, respiratory, infection, pathology, and virology. The Corpus of English Journal Articles is composed of high-impact journal articles (CEJAHI) with highly cited times ($\geqslant$ 300) published in the journals with high division (Q1/Q2) and high impact factor ($\geqslant$ 10), as well as moderate-impact journal articles (CEJAMI) with moderately cited times ($\leqslant$ 10) published in the journals with moderate division (Q3/Q4) and moderate impact factor ($\leqslant$ 1).

| Subcorpora | Pages | Tokens | Types | Journal Num. | Average: Article/Journal |
|---|---|---|---|---|---|
| CEJAHI | 100 | 318,848 | 12,668 | 43 | 2.33 |
| CEJAMI | 100 | 302,439 | 16,813 | 67 | 1.49 |

Table 1: Basic Information of CJAHI and CJAMI

## 3.2 Preprocessing of Dataset

This study only retains complete sentences in our corpus. The citation within the text, references, figures, tables, funding, formulas, acknowledgments, appendices, author and journal information, non-English characters that are not relevant to the overall statement of the text were temporarily deleted. However, it should be noted that some numbers and author names are used as grammatical components in the sentences. To not damage the fluency of the text, we reserved the numbers and author names that serve as grammatical components in the main sentence. Therefore, the corpus used for further analysis of our study only contains the title, the abstract, keywords, the main body of the article, summary box, and numbers and author names that serve as grammatical components. The retained texts were not lower cased and punctuation removed because the text analyzer needs to retain the original text.

## 3.3 Source of Feature Sets

The feature sets of our sudy come from two widely used text analyzers and dictionaries. Coh-Metrix 3.0 (Graesser et al., 2003; Graesser et al., 2004) is by far one of the most widely used analyzers for written texts due to its varieties of linguistic and textual indicators (e.g., Graesser et al., 2011; McNamara et al., 2014; Tegge & Parry, 2020). It contains 11 major grammatical categories (108 indicators) involving descriptive, text easability principal component scores, referential cohesion, latent semantic analysis (LSA), lexical diversity, connectives, situation mode, syntactic complexity, syntactic pattern density, word information, and readability. Coh-Metrix has been often used to explore the use of vocabulary, syntax, and semantics in the students' English writings (Burgess, Livesay, & Lund, 1998; Graesser, McNamara, & Louwerse, 2003), while our study uses it for analyzing the grammatical features of journal articles with binary academic impacts.

We adopted LIWC 2015 (Linguistic Inquiry Word Count) to analyze linguistic uses at the rhetoric level, which we also consider as micro-level linguistic features. This dictionary can calculate the proportion of words used in the text involving emotion, medical treatment, health, cognitive psychology, thought style, social attention, power, and ideology (Smirnova et al., 2017). The software has undergone a rigorous psychometric evaluation (Pennebaker et al., 2015) through the intuitive evaluation of native speakers and the retrieval of the actual corpus, so it can effectively and reliably reflect the potentially socio-cognitive word categories. LIWC 2015 uses two variables to quantify the text, namely "Summary Variables" and "Linguistic Dimensions". "Summary Variable" is composed of analytical thinking, clout, authenticity, and emotional Tone, which can calculate the degree of words related to logical thinking and narrative thinking (Pennebaker et al., 2014), professionalism, and heuristics (Kacewicz et al., 2014), personal and impersonal (Newman et al., 2003), positive and negative emotions (Cohn et al., 2004). On the other hand, "Linguistic Dimensions" mainly includes words related to emotion, cognition, perception, and psychology. LIWC has been majorly employed in the text analysis of news articles (e.g., Tay, 2021) and patient reports (e.g., Tay, 2017), and its application in academic text analysis has not yet been used in the previous studies. This article attempts to investigate its applicability in the recognition of linguistic features of journal articles.

Through an initial analysis of the linguistic indicators provided by Coh-Metrix 3.0 and LIWC 2015, we retained 24 categories of 146-dimensional linguistic indicators for further analysis.

## 3.4 Feature Engineering

In general, feature engineering converts raw data into more efficient encoding methods, which is easy to find high-quality features. High-quality features can help researchers choose simple models to make calculations faster, results easier to interpretation, and models more convenient to maintainence (Li, 2019). Given that the initial feature sets of this study have 146 dimensions in total, feature engineering can be used to reduce their dimensionality and make good use of features. , There are three steps in the feature engineering: feature brainstorm, feature selection, and feature monitor (Li et al., 2019).

### 3.4.1 Feature Brainstorm

Feature brainstorm requires a complete understanding of research questions and the mastery of the data background, combined with professional domain knowledge, mathematical knowledge, and even experience and intuition. As the above mentions, this study chose Coh-Metrix 3.0 and LIWC 2015 to initiate feature sets.

### 3.4.2 Feature Selection

Feature selection is the core of feature engineering, whose purpose is to reduce noise and redundancy by trimming features (Wang et al., 2017). Choosing a suitable feature subset can reduce the interaction between features and avoid overfitting/underfitting to improve the model's generalization ability and computing speed; on the other hand, it can improve the interpretability of the model (Zhao & Zhang, 2018). There are three types of commonly used methods in feature selection: filtering, wrapping, and embedding (Kuhn & Johnson, 2018).

(1) Filtering: Select features that are highly correlated with the response variable, less correlated with other features, and easy to recognize by themselves.

(2) Wrapping: Continuously input different feature subsets into the model, and recursively select or exclude features according to the classification effect.

(3) Embedding: Use a machine-learning algorithm to train the model to obtain the weight coefficient of each feature, and select the feature(s) according to the weight.

This paper uses the filtering method to select features (Chi-square test, Mann-Whitney *U* Test, Information Gain, ReliefF algorithm). Since the filtering method does not need to build a classifier, it helps save time. In addition, it is an optional framework for processing high-dimensional small-sample data (Li et al., 2019).

### 3.4.3 Feature Monitor

Feature monitor is usually combined with feature selection and model training and testing: feature selection initially extracts compelling features. The model's performance on selected features was compared with that on original feature sets to verify whether the prediction of the selected features is accurate and consistent (Xing et al., 2016).

So the next part briefly describes the mathematical principles in the supervised machine learning models used by this study.

### 3.5 Machine Learning Model

The data set has two clear labels of "high impact" and "moderate impact", so this study is a binary classification task, and supervised learning models are better to complete such a task (Hoskins & Hooff, 2005; Al-Shabandar et al., 2018). In order to better solve the problem of data overfitting and underfitting, this study uses four classic machine learning classification models: Ridge Regression, Lasso Regression, Decision Tree, and Random Forest.

### 3.5.1 Ridge and Lasso Regression

Since ridge and lasso regression both belong to the regression model, and the theory is similar, they are introduced together. Regression is a type of supervised learning. It uses a line/curve to fit the known function and predict the unknown function, i.e., the relationship between input and output variables. Considering any data has noise which might lead to overfitting or underfitting issues, ridge and lasso regularization emerge. They increase the loss function of the regularization term by adding a penalty factor to linear regression parameters to prevent overfitting or underfitting. The addition of lasso or ridge regularization can make the weight as small as possible, and finally, construct a model with all parameters relatively small. It is generally believed that models with small parameter values can adapt to different data sets and also avoid overfitting or underfitting to a certain extent (CSDN, 2019).

Ridge regression is the addition of L2 regularization to the standard linear regression, as shown on the right side of the plus sign in the following formula:

$$J = \frac{1}{2}\sum_{i=1}^{m}(h_\theta(x^{(i)}) - y^{(i)})^2 + \lambda \sum_{j=1}^{n}\theta_j^2$$

L2 regularization makes the parameters smaller and slows down the phenomenon of overfitting or underfitting.

In the same way, lasso regression is to add L1 regularization based on standard linear regression, as shown on the right side of the plus sign in the following formula:

$$J = \frac{1}{2}\sum_{i=1}^{m}(h_\theta(x^{(i)}) - y^{(i)})^2 + \lambda ||w||_1$$

Under the L1 model, the input features with no (or negligible) contribution from the regression equation are automatically deleted to achieve model optimization.

### 3.5.2 Decision Trees

Unlike ridge and lasso regression, a decision tree is a non-linear model. It is one of the most practical and widely used inductive reasoning methods. Its processing flowchart is tree-shaped, followed by the *if-then* rule (as shown in Figure 2).

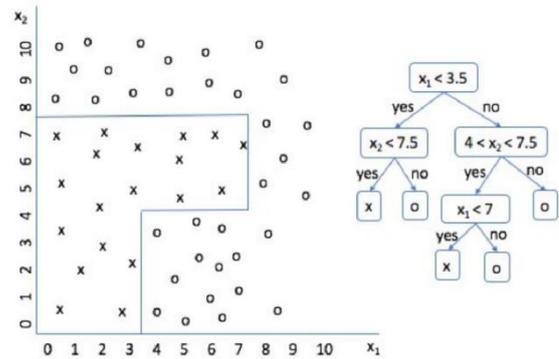

Figure 2: Flowchart of Decision Tree Classification (from the internet)

As Figure 2 indicates, a decision tree is mainly generated according to the following three steps (Zhang, 2020):

(1) Feature selection: Choose one feature from multiple features.

(2) Decision tree generation: According to the selected features above, the child nodes are determined sequentially from top-down. Only when samples belonging to the same category are left in the data set, the growth of the decision tree is stopped.

(3) Pruning: The decision tree is prone to over-fitting problems during the generation process, usually solved by pruning, including two pruning techniques: pre-pruning and post-pruning (Brownlee, 2019).

### 3.5.3 Random Forests

Since the emergence of the random forest model, it has been rapidly developed and widely used in various fields such as medicine, architecture, economics, and aviation (e.g., Wang, 2016; Pei, 2018; Wang & Xia, 2018). Generally, such an algorithm performs well on the data set, has fewer parameters to be adjusted, and has no overfitting problems; at the same time, it can also efficiently process high-dimensional data and have better anti-noise capabilities (Liu, 2016).

The mechanisms behind the random forest can be summarized into the following stages (Li, 2019): firstly, it is formed by combining a large number of single classification and regression trees. Each tree is based on a randomly independent vector, and bagging and random sampling (a classical voting method) are used to obtain the final classification result (Figure 3). The classification efficiency of a single tree and the correlation between the classification trees directly affects the generalization error of the entire random forest. The classification efficiency is defined as the accuracy of the classification tree to classify new test data (Brownlee, 2020).

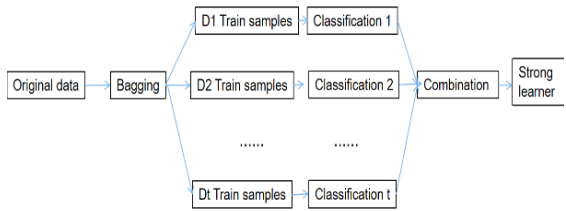

Figure 3: Flowchart of Random Forest

## 4  Result of Feature Selection

According to four filtering methods in the feature selection (Chi-Square test, Mann-Whitney *U* test, information gain, and ReliefF), we received 25 features including imageability for content words, third-person plural pronouns, content word overlap of all sentences, content word overlap of adjacent sentences, LSA overlap of adjacent sentences, LSA overlap of adjacent sentences, LSA given/new of sentences, noun overlap of all sentences, word concreteness, concreteness for content words, second-person pronouns, argument overlap of all sentences, noun overlap of adjacent sentences, CELEX log frequency for all words, referential cohesion, anaphor overlap of adjacent sentences, auxiliary words, focus on past, focus on present, focus on future, article, number, positive emotion, causative words.

Next, these features were all put into the machine learning model to test the consistency and accuracy of distinguishing high- and moderate-impact journal articles.

## 5  Result of Feature Prediction

### 5.1 Data Partition

The labeled data set was divided into three parts: the training set, the validation set, and the test set. The training set is suitable for the data samples for model fitting, and the validation set is a separate set of training samples to adjust the model's hyperparameters and evaluate the model. The test set is used to evaluate the generalization ability of the final model (Kloft et al., 2014). The sampling method is bootstrapping and k-fold cross-validation.

The process of dividing the data set in this study is as follows:

(1) Divide the original labeled data set into two parts according to the ratio of 8:2; 80% of the data is used for model training and validation, and the remaining 20% is for model testing.
(2) Use 80% of the data set according to bootstrapping for model training.
(3) Use k-fold cross-validation to adjust the model parameters. K is set as 10 in our study.
(4) 20% of the data set is used to judge the performance of the model.

The critical point in data partition is: the data is seriously imbalanced, so it is necessary to ensure the consistency of the distribution of each data subset to be divided. This research uses the *StratifiedKFold* function in the machine learning package *sklearn* to satisfy this requirement. On the other hand, since the imbalance of the data is the inherent trait of the data, it does not consider the use of undersampling, oversampling, SMOTE, and other methods to eliminate such imbalance. None of the models in this study need normalizing the scale of feature sets.

### 5.2 Model Performance

Four supervised learning models were used to classify English journal articles with different impacts based on these 25 features.

| Models | AUC | Classification Accuracy | F1 | Precision | Recall |
|---|---|---|---|---|---|

| | | | | |
|---|---|---|---|---|
| Random forest | 94.5% | 90.5% | 90.5% | 90.7% |
| Decision tree | 90.6% | 89.0% | 89.0% | 89.1% |
| Ridged regularization | 87.3% | 78.5% | 78.5% | 78.6% |
| Lasso regularization | 85.7% | 79.0% | 79.0% | 79.1% |

Table 2: Classification results of four machine learning models (10-fold cross-validation, stratified)

Table 2 shows the fine-tuning results of the four models under the bootstrap sampling method, and the 10-fold cross-validation is selected. The closer the value (AUC, classification accuracy, F1, precision, and recall) is to 100%, the stronger the classification ability of the model is. The random forest model outperforms the other models with AUC 94.5%, exceeding the other model performance.

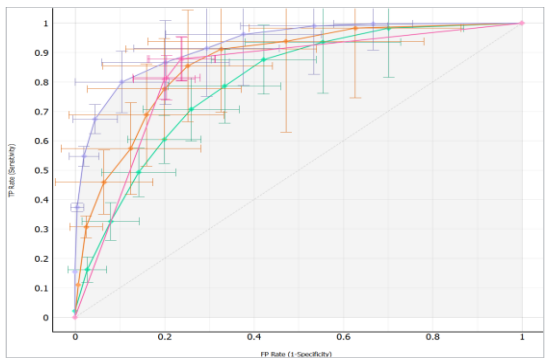
Figure 4a: ROC analysis of CEJAHI as DV

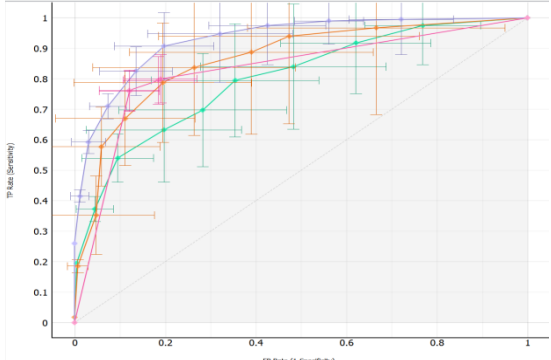
Figure 4b: ROC analysis of CEJAMI as DV
Note: purple: random forest; green: ridge regularization; pink: decision trees; yellow: lasso regularization
Figure 4: ROC analysis of four machine learning models

Furthermore, Figure 4 can prove a better performance of the random forest. It is a ROC analysis: Figure 4a describes the relationship of the mean true-positive rate of high-impact labels classified as high-impact labels (*y*-axis) and high-impact labels classified as moderate-impact labels (*x*-axis); Figure 4b the relationship of the mean true-positive rate of moderate-impact labels classified as moderate-impact labels (*y*-axis) and moderate-impact labels classified as high-impact labels (*x*-axis). The area between the x-axis and the line is more extensive, the better the classification effect of the model is. In general, the random forest has a better performance. The Confusion Matrix in Table 3 also specifies the classification accuracy of the random forest with a satisfactory performance. As Brownlee (2019, 2020) said, the random forest model based on the Bootstrap Sampling method is a robust classifier. Our research also proves this point.

| | | Predicted values | |
|---|---|---|---|
| | | CEJAHI | CEJAMI |
| Observed values | CEJAHI | 88.2% | 6.7% |
| | CEJAMI | 11.8% | 93.3% |

Table 3: Confusion Matrix based on random forest

### 5.3 Feature Performance

Since the random forest model is better, we would then verify the predictive power of these 25 features.

Figure 5 presents the feature importance under the random forest model.

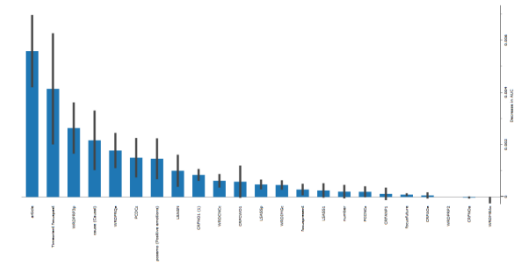
Figure 5: Feature Importance in Random Forest

The 24 features proved their significance to interpret the journal articles with different impacts. Compared with the random forest model trained with all 146 indicators (AUC = 95.2%, CA = 88.1%, F1 = 88.1%, Precision = 88.1%, Recall = 88.1%), the performance of the random forest model trained with these 24 indicators outperforms its CA, F1, precision, recall, though not as high as its AUC. Similar or better performance on 24 features indicates that the 24 indicators have a good prediction.

Figure 6 visualizes the classification results of the random forest model by Pythagorean forest, which shows all learned decision tree models from the Random Forest. It displays them as Pythagorean trees, each visualization pertaining to one randomly constructed tree. The best tree has the shortest and most strongly colored branches, meaning that few attributes split the branches well. As Figure 6 demonstrates, there are 20 trees constructed by the random forest algorithm. To reduce the case of too-many fluctuations, we logarithmically transformed the data. Blue represents the label of high-impact journal articles, whereas red represents the label of moderate-impact journal articles. According to the above criterion, the best tree is the left top.

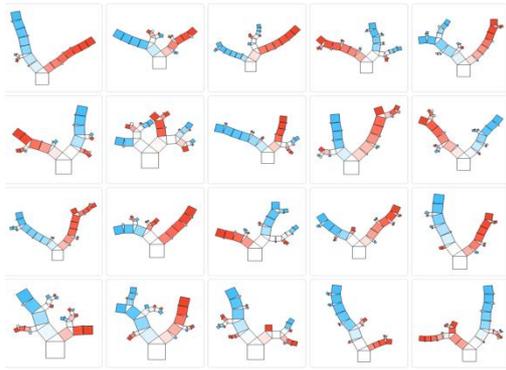
Figure 6: Pythagorean forest

Moreover, we displayed the weights of features in the tress by Pythagorean Tree as Figure 7 shown.

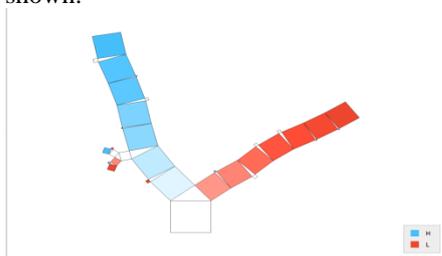
Figure 7: Pythagorean Tree on features

There are 33 nodes and the depth is 7. The most important features that can better interpret high-impact journal articles are noun overlap in all sentences $> 0.230$, cause $\leqslant 2.280$, focusfuture $\leqslant 1.035$, focuspast $> 2.650$, posemo $> 0.460$, LSA given/new $\in (0.245, 0.415)$, while the features that can better interpret moderate-impact journal articles are Text Easability PC Connectivity $> -3.321$, Text Easability PC Referential cohesion $\leqslant 59.315$, Third person plural pronoun incidence $> 0.697$, focuspast $\leqslant 4.615$, number $> 3.230$, cause $\in (2,280, 4.515)$.

## 6 Conclusion

In summary, we used four feature selection methods to reduce the feature dimensions. We verified the selected 24 features, including word imagery, third-person plural pronouns, all sentence actual word overlap, adjacent sentence actual word overlap, the semantic overlap of all sentences, the semantic overlap of adjacent sentences, the ratio of old information to new information in all sentences, the overlap of all sentence nouns, text ease of vocabulary actuality, content word meaning, content word concretization, all sentence argument overlap, adjacent sentence noun overlap, modal verb, focus past, focus present, focus future, articles, numbers, positive emotions, causal vocabulary can better distinguish journal articles with different impacts.

Besides, we found that based on these 24 language features, the random forest model can be more consistent and accurate to classify high- and moderate-impact journal articles. The selected linguistic features at the micro-level and machine learning model with better performance can provide reference for future automatic analysis and feedback systems of English research articles.